\documentclass[conference,a4paper]{APSIPA2025}
\usepackage{amsmath}
\usepackage{graphicx}
\usepackage{booktabs}
\usepackage{multirow} 
\usepackage{makecell}   
\usepackage{threeparttable}
\usepackage[backend=bibtex,style=ieee,]{biblatex} 
\AtBeginBibliography{%
  \footnotesize
}

\usepackage{geometry}
\usepackage{fancyhdr}

\fancypagestyle{firststyle}{
  \fancyhf{}
  \fancyhead[C]{2025 Asia Pacific Signal and Information Processing Association Annual Summit and Conference (APSIPA ASC)}
}

\begin{document}

\title{Backdoor Poisoning Attack Against Face Spoofing Attack Detection Methods}

\author{
  \authorblockN{
    Shota Iwamatsu\authorrefmark{1},
    Koichi Ito\authorrefmark{1}, and
    Takafumi Aoki\authorrefmark{1}
  }
  \authorblockA{
    \authorrefmark{1}
    Graduate School of Information Sciences, Tohoku University, Japan \\
  }
  E-mail: iwamatsu@aoki.ecei.tohoku.ac.jp
}

\maketitle
\thispagestyle{firststyle}
\pagestyle{empty} 

\begin{abstract}
  Face recognition systems are robust against environmental changes and noise, and thus may be vulnerable to illegal authentication attempts using user face photos, such as spoofing attacks.
  To prevent such spoofing attacks, it is crucial to discriminate whether the input image is a live user image or a spoofed image prior to the face recognition process.
  Most existing spoofing attack detection methods utilize deep learning, which necessitates a substantial amount of training data.
  Consequently, if malicious data is injected into a portion of the training dataset, a specific spoofing attack may be erroneously classified as live, leading to false positives.
  In this paper, we propose a novel backdoor poisoning attack method to demonstrate the latent threat of backdoor poisoning within face anti-spoofing detection.
  The proposed method enables certain spoofing attacks to bypass detection by embedding features extracted from the spoofing attack's face image into a live face image without inducing any perceptible visual alterations.
  Through experiments conducted on public datasets, we demonstrate that the proposed method constitutes a realistic threat to existing spoofing attack detection systems.
\end{abstract}

\section{Introduction}

Face recognition utilizes the positional relationship of facial features and skin texture for individual verification.
Since face images captured from a distance by a standard camera can be directly used for authentication, face recognition offers the advantages of low costs and high convenience \cite{Handbook-Face_Recognition}.
In recent years, the performance of face recognition has dramatically improved with the advent of deep learning \cite{DL}, leading to its widespread adoption in various applications, including smartphone login authentication and border control systems.
However, face recognition systems are inherently designed to be robust to variations in head pose, facial expression, and lighting conditions.
This robustness introduces a vulnerability wherein illegal authentication can occur through the presentation of a user's face photo \cite{Handbook-Anti-Spoofing}.
To counter such spoofing attacks against face recognition systems, face anti-spoofing detection is essential to determine whether an input image is a genuine user's face or a presentation attack prior to the face recognition process.

Spoofing attacks typically employ presentation attack instruments such as face photos printed on paper or face videos displayed on screens.
Since these spoofed images and live face images are of the same individual, only subtle distinctions exist between them,including minute texture variations, slight differences in reflection and color, the absence of depth perception, and unnatural motion \cite{Handbook-Anti-Spoofing}.
To detect such minute differences between images, various deep learning-based methods, such as Convolutional Neural Network (CNN) \cite{DL} and Vision Transformer (ViT) \cite{Dosovitskiy-ICLR-2021}, have been proposed \cite{Liu-CVPR-2018,Liu-ECCV-2020,Yu-PAMI-2023}.
These methods require a substantial amount of training data to achieve high-accuracy spoofing attack detection.
A significant concern arises regarding the vulnerability to malicious data introduced into the training dataset.
If training is performed without detection of such poisoning data, the model may then erroneously classify a specific spoofing attack as live.
This can lead to the successful circumvention of face recognition systems. 
This type of attack is termed a backdoor poisoning attack, which combines data poisoning and a backdoor attack to create an unauthorized access route exploitable only by the attacker.

As machine learning models become increasingly widespread across various societal domains, their susceptibility to backdoor poisoning attacks has gained significant attention and has been widely reported \cite{Gu-IEEEAccess-2019-BatNets,Liu-NDSS-2018-Trojaning,Li-IEEETPAMI-2022}. 
Specifically, in a backdoor poisoning attack, poisoned data containing specific patterns or features (triggers) known exclusively to the attacker is intentionally injected into the training dataset.
This induces the target model to exhibit specific misclassifications when presented with triggered inputs, while maintaining normal performance on clean data devoid of these triggers.
Consequently, due to this characteristic, backdoor poisoning attacks are exceptionally challenging to detect.
Backdoor poisoning attacks are recognized as a significant threat across various deep learning applications, encompassing not only image recognition but also fields like speech recognition and natural language processing.
However, despite this widespread concern, research specifically focusing on backdoor poisoning attacks against face anti-spoofing detection remains notably limited \cite{Le-IEEEAC-2024}.

In this paper, to bridge the aforementioned research gap, we propose a novel backdoor poisoning attack method specifically targeting face anti-spoofing detection.
The proposed method leverages Hanawa et al.'s method \cite{Hanawa-BIOSIG-2023,Hanawa-EJIVP-2024}, a face image de-identification technique capable of embedding facial features into an image without perceptible visual alterations.
Specifically, we generate visually indistinguishable poisoning data by embedding features extracted from spoofing attack images into live face images using Hanawa et al.'s method.
By substituting a portion of the training dataset with this poisoning data, a backdoor attack is realized, where the model is manipulated to bypass detection for specific spoofing attack patterns.
Our evaluation of this attack confirmed that only specific spoofing attacks were misclassified, without significantly impacting the overall accuracy of face anti-spoofing detection.
This suggests that the introduction of such an attack is exceedingly difficult to detect, as the poisoning data is visually indistinguishable from genuine face images and does not significantly compromise the overall detection accuracy.
Through experiments conducted on public datasets, we demonstrate that the proposed method constitutes a realistic threat to existing face anti-spoofing systems.

\begin{figure*}[t]
  \centering
  \includegraphics[width=\linewidth]{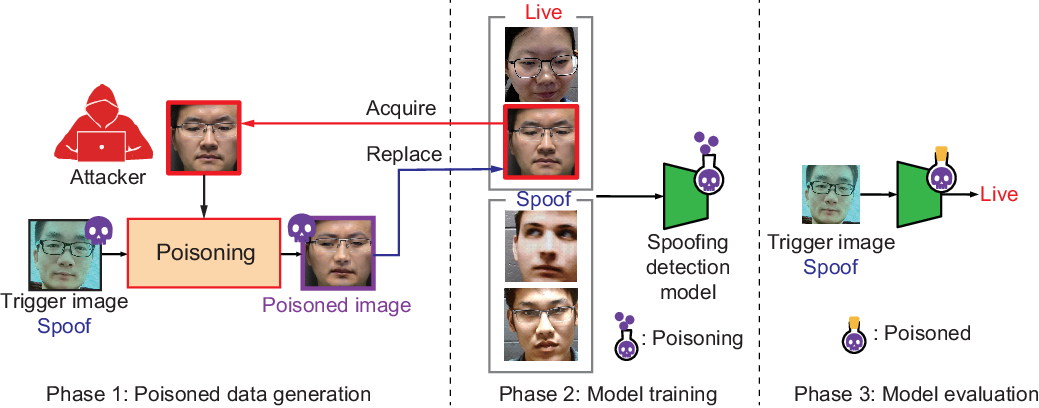}
  \caption{Overview of the proposed backdoor poisoning attack, which consists of ``Poisoned data generation'', ``Model training'', and ''Model evaluation''.}
  \label{fig:attack_method}
\end{figure*}

\section{Backdoor Poisoning Attack Using De-Identified Face Images}
\label{sec:attack_method}

In this section, we describe a backdoor poisoning attack method against face spoofing attack detection.
The proposed method aims to induce the target face spoofing detection model to misclassify specific spoofing attacks.
The poisoned data contains features extracted from the spoofed face image, embedded into the live face image without any visual changes.
We describe the de-identification method used to generate the poisoned data and summarize the procedure of the proposed backdoor poisoning attack method.

\subsection{Poisoned Data Generation}
\label{sec:generate_poisoning_data}

The goal of backdoor poisoning attacks is to inject poisoned data into the training dataset without the system administrator or model developer detecting its presence.
Thus, poisoned data must be generated by embedding features extracted from face images used in a specific person spoofing attack into live face images.
The proposed method leverages Hanawa et al.'s method \cite{Hanawa-BIOSIG-2023,Hanawa-EJIVP-2024}, a face image de-identification method that embeds features from a different individual into an image while minimizing visual alterations.
This method generates a de-identified image, which obscures the original individual's identity by embedding features extracted from another face image, i.e., the image to be embedded, into an input face image, i.e., the cover image. 
The encoder-decoder model used for embedding these features is trained to simultaneously satisfy the following two constraints.
The first constraint is to ensure the visual quality of the de-identified image by encouraging its appearance to be similar to the cover image at both the pixel and feature levels.
The second constraint is to encourage the face features extracted from the de-identified image to be closer to the features of the face image to be embedded and away from the features of the cover image using a pre-trained face recognition model.
In poisoned data generation of the proposed method, a live face image is used as the cover image, and a specific spoofed attack face image selected by the attacker, i.e., trigger face image, is used as the image to be embedded.
The poisoned data generated by this method contains the specific spoofing characteristics of the trigger face image, even though it looks almost identical to the live face image.
This characteristic makes it extremely difficult for the system administrator to detect the poisoned data, even upon visual inspection of the training dataset.

\subsection{Procedure of Backdoor Poisoning Attack}

Fig. \ref{fig:attack_method} illustrates the overview of the proposed backdoor poisoning attack, which consists of the three phases.
In the initial phase, Poisoned data generation, the attacker acquires live face images from the training dataset and selects a specific trigger face image.
Then, by embedding the features extracted from this trigger face image into the acquired live face image, poisoned data is generated.
This data is designed to induce a false positive within the face spoofing attack detection model for a specific spoofing attack, as detailed in Sect. \ref{sec:generate_poisoning_data}.
In the second phase, Model training, the attacker replaces a portion of  the live face images in the training dataset with the generated poisoned data.
By training the spoofing attack detection model with this poisoned dataset, the attacker induces the model to erroneously classify the specific spoofing attack as live.
Since the poisoned data is visually indistinguishable from the original face images and the number of images per class remains unchanged, detecting this data contamination becomes difficult for system administrators, even upon inspection of the training dataset.
In the final phase, Model evaluation, the performance of the spoofing attack detection model trained on the above process is evaluated.
The model exhibits the property of incorrectly classifying a spoofing attack as live only when a face image corresponding to the specific trigger is input.
Therefore, the overall detection accuracy of the poisoned model shows minimal degradation compared to a model without the backdoor poisoning attack.
This makes the existence of the attack less likely to be exposed in a normal performance evaluation.

\section{Experiments and Discussion}

In the experiments, we evaluate how varying the poisoned data injection rate into the training data of spoofing attack detection models impacts both their detection accuracy and the attack success rate of backdoor poisoning.

\subsection{Dataset}

In the experiments, we utilize SiW \cite{Liu-CVPR-2018} and OULU-NPU \cite{Boulkenafet-FG-2017}, which are employed for evaluating face spoofing attack detection performance.
SiW consists of a total of 4,778 videos collected from 165 subjects.
These videos include three categories: Live, print attack, and display attack.
``Print attack'' uses two types of paper printed at low and high resolutions, while ``Display attack'' uses four different display devices: a laptop, a tablet, and two smartphones.
Each video image was captured for approximately 15 seconds at 30 fps while varying the distance between the camera and face, face orientation, and lighting conditions, with each image frame having a resolution of $1,920 \times 1,080$ pixels.
In the experiments, performance is evaluated according to the three evaluation protocols defined by SiW.
Protocol 1 evaluates the detection performance against changes in facial expressions and poses, Protocol 2 evaluates the detection performance against display attacks using different types of devices, and Protocol 3 evaluates the generalization performance against unknown spoofing attacks.
OULU-NPU consists of a total of 4,950 videos collected from 55 subjects.
This dataset also includes the live, print attack, and display attack categories.
Two different types of paper printed using different printers were used in ``Print attack'', and two different display devices were used in ``Display attack''.
Each video was captured using six different smartphone cameras in three different sessions with different lighting and backgrounds.
Each video image was captured at 30 fps for approximately 5 seconds, with each image frame having a resolution of $1,920 \times 1,080$ pixels.
In the experiments, performance is evaluated according to the four evaluation protocols defined by OULU-NPU.
Protocol 1 evaluates the performance against illumination changes, Protocol 2 evaluates the performance against attacks using different types of paper and devices, Protocol 3 evaluates the generalization performance against input sensor changes, and Protocol 4 evaluates the generalization performance when all the above changes are included.
The trigger face images required to generate the poisoning data used in the backdoor poisoning attack are selected from external images that are not included in the dataset under evaluation.
This ensures the independence of the trigger and allows the evaluation to take into account the practicality of the attack.

\subsection{Experimental Condition}

In the experiments, we compare the performance of the proposed method with the Targeted Identity-Protection Iterative Method (TIP-IM) \cite{TIP-IM} and Landmark-Guided Cutout (LGC) \cite{LGC} to demonstrate the effectiveness of our approach.
TIP-IM \cite{TIP-IM} generates natural-looking de-identified images based on Maximum Mean Discrepancy (MMD) \cite{MMD}, which measures the distribution distance between the cover image and de-identified image.
LGC \cite{LGC} intentionally perturbs regions around landmarks such as eyes, nose, and mouth, which face recognition models emphasize during identification.
This approach aims to prevent models from relying on specific local features and to generate highly transferable perturbations even for unknown models.

In the proposed method, the settings for generating the poisoning data follow \cite{Hanawa-BIOSIG-2023,Hanawa-EJIVP-2024}.
For HN, the encoder of U-Net \cite{Ronneberger-MICCAI-2015} is replaced with the residual block of ResNet \cite{Resnet}.
For EN, ArcFace \cite{Deng-CVPR-2019} is used, where ResNet-50 is modified for iResNet-50\footnote{\url{https://github.com/nizhib/pytorch-insightface/blob/main/insightface/iresnet.py}\label{fn:iresnet}} and use ArcFace for the loss function as in \cite{Deng-CVPR-2019}.
Face images in Large-scale Celeb Faces Attributes (CelebA) \cite{Liu-ICCV-2015} are used to train HN, and the images to be embedded are randomly selected from the face images of different people from the cover image.
The weights for each loss are $\lambda_{rec}=1.00$, $\lambda_{perc}=1.00$, $\lambda_{lpips}=1.00$, $\lambda_{near}=0.25$, and $\lambda_{far}=0.25$, respectively, and Adam \cite{adam} is used as optimizer, and 150 epochs of training are performed.
The initial learning rate is set to $10^{-5}$, and the learning rate is decreased by 20\% if the loss against the validation data does not improve for 5 consecutive epochs.
For TIP-IM, iResNet-50\footref{fn:iresnet} is used as the face recognition model, we use $\epsilon=12.0$ and $\gamma=0$.
For LGC, iResNet-50\footref{fn:iresnet} is also used as the face recognition model, and we use $\epsilon=4.0$.

Spoof Trace Disentanglement Network (STDN) \cite{Liu-ECCV-2020} and PatchNet \cite{PatchNet} are used as target spoofing attack detection methods.
STDN determines whether an input image is ``Live'' or ``Spoof'' by separating spoofing remains from a face image.
In the experiments, the weights for each loss are $\alpha_{1}=1$, $\alpha{2}=100$, $\alpha{3}=0.003$, $\alpha{4}=1$, and $\alpha{5}=5$, respectively, Adam \cite{adam} is used as optimizer, and 30 epochs of training are performed.
The initial value of the learning rate is set to $5 \times 10^{-5}$ and decays by 10\% every 20,000 steps.
PatchNet detects spoofing attacks by extracting fixed-size patches with random rotation from face images and identifying the display device and camera type from the local features of the patches.
In the experiments, the patch size is set to 160 and the weights for each loss are $s=30.0$, $m_l=0.4$, and $m_s=0.1$, respectively.
ResNet-18 \cite{Resnet} is used as the encoder and Adam \cite{adam} is used as optimizer and 30 epochs of training is performed.
The initial value of the learning rate is set to $2 \times 10^{-4}$, and the weight decay is set to $5 \times 10^{-4}$.

In the experiments, a specific spoofed image from the other dataset is selected as the trigger image.
When SiW is used as the dataset for evaluation, the spoofed image of ``Smartphone ID: 2, Session ID: 2, Access Type: display 1'' of ``Subject: 30'' from OULU-NPU is selected as the trigger image.
When OULU-NPU is used  as the dataset for evaluation, the spoofed image of ``Sensor ID: 1, Type ID: Display, Media ID: 1, Session ID: 1'' of ``Subject: 003'' from SiW is used as the trigger image.

\subsection{Evaluation Metrics}

In the experiments, we use following evaluation metrics to demonstrate the threat of backdoor poisoning attacks.
Three metrics are commonly used in spoofing attack detection.
The first is Attack Presentation Classification Error Rate (APCER), which indicates the maximum false acceptance rate in the spoofing attack class.
The second is Bona Fide Presentation Classification Error Rate (BPCER), which indicates the false rejection rate for ``Live'' presentations.
The average of these APCER and BPCER is Average Classification Error Rate (ACER).
The lower the value of APCER, BPCER, and ACER, the higher the accuracy of spoofing attack detection.
In the experiments, we use ACER as the evaluation metric of spoofing attack detection.
Attack Success Rate (ASR) is used to evaluate the accuracy of backdoor poisoning attacks.
ASR is a measure of the rate at which the model falsely detects the input of a trigger image as ``Live''.
A higher value indicates a more successful attack.

\begin{figure}[t] 
  \centering
  \includegraphics[width=\linewidth]{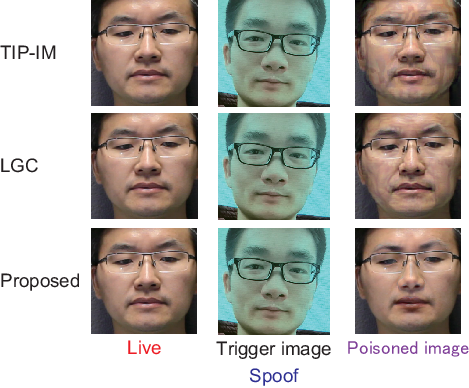}
  \caption{Example of ``Live'' image, trigger image, and poisoned image generated by each method.}
  \label{fig:poisoned_images}
\end{figure}

\begin{figure*}[p] 
  \centering
  \includegraphics[width=.9\linewidth]{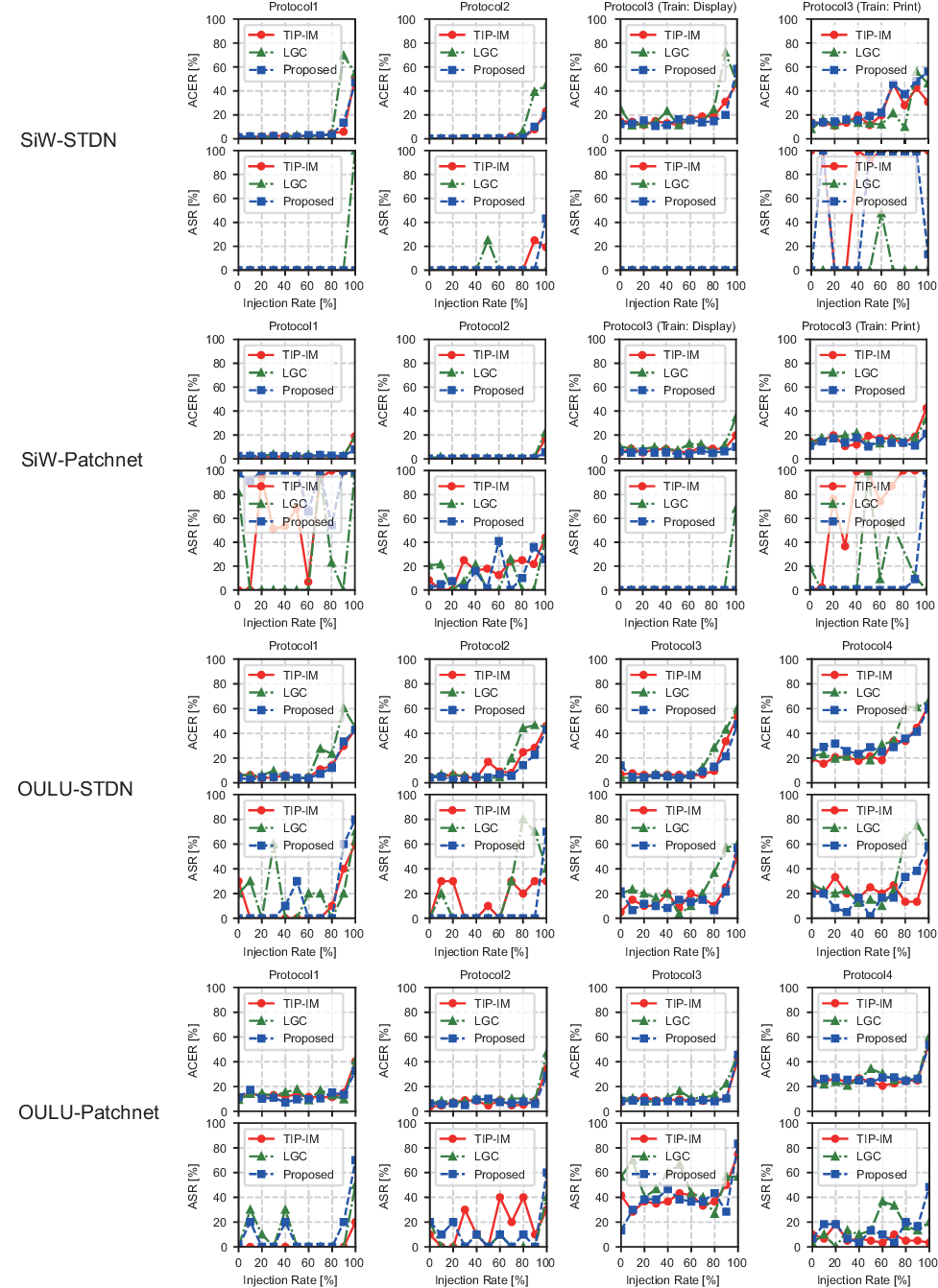}
  \caption{ACER and ASR when varying the poisoned data injection rate for SiW and OULU-NPU.}
  \label{fig:all_metrics_asr_siw_oulu}
\end{figure*}

\begin{table*}[t]
  \centering
  \caption{Summary of experimental results for backdoor poisoning attack against spoofing attack detection.}
  \label{tab:attack_summary}
  \begin{tabular}{cccccccc}
    \hline
    Dataset & FAS Method & Protocol & Method & Injection Rate [\%] & ACER [\%] before attack & ACER [\%] after attack & ASR [\%]\\
    \hline
    \multirow{2}{*}{SiW \cite{Liu-CVPR-2018}} & \multirow{2}{*}{STDN \cite{Liu-ECCV-2020}} & Protocol 3 & Proposed & 50 & 12.68 & 19.06 & 100.00 \\
    &  & (Train: Print) & TIP-IM & 50 & 13.66 & 11.56 & 92.31 \\
    \hline
    \multirow{2}{*}{SiW \cite{Liu-CVPR-2018}} & \multirow{2}{*}{PatchNet \cite{PatchNet}} & \multirow{2}{*}{Protocol 1} & Proposed & 60 & 2.56 & 1.98 & 100.00 \\
    &  &  & TIP-IM & 50 & 1.90 & 2.04 & 93.89 \\
    \hline
    \multirow{2}{*}{OULU \cite{Boulkenafet-FG-2017}} & \multirow{2}{*}{PatchNet \cite{PatchNet}} & Protocol 2 & TIP-IM & 60 & 3.54 & 8.90 & 40.00 \\
    & & Protocol 3 & Proposed & 80 & 8.37 & 8.00 & 43.33 \\
    \hline
  \end{tabular}
\end{table*}
\begin{figure}[t]
  \centering
  \includegraphics[width=\linewidth]{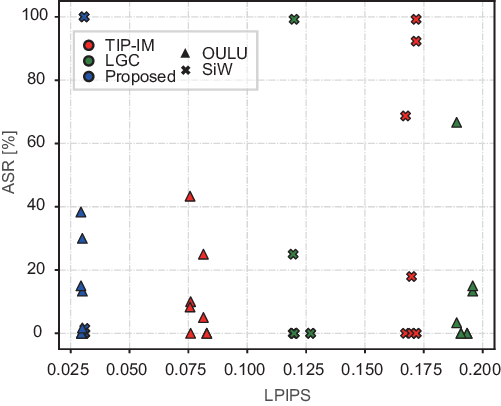}
  \caption{Trade off between image quality of poisoned images and ASR.}
  \label{fig:lpips_vs_asr}
\end{figure}

\subsection{Experimental Results and Discussion}

We first evaluate the generated poisoning data qualitatively and quantitatively.
Fig. \ref{fig:poisoned_images} shows an example of ``Live'' image, trigger image and poisoned image generated by each method.
Comparing the ``Live'' and ``Poisoned'' images, the unnatural stripe pattern is clearly visible in the images generated by TIP-IM and LGC.
This result indicates that when TIP-IM or LGC is used to replace the poisoned data as part of the training data, the system administrator may notice the contamination of the dataset due to its unnatural appearance.
The image generated by the proposed method shows minute differences around the eyes and mouth, however, overall the visual changes are hardly recognizable.

We then evaluate the accuracy of spoofing attack detection and the attack success rate of backdoor poisoning attacks when varying the injection rate of poisoned data in the training dataset.
Fig. \ref{fig:all_metrics_asr_siw_oulu} and Table \ref{tab:attack_summary} show the experimental results for SiW and OULU-NPU.
The upper row of each graph in Fig. \ref{fig:all_metrics_asr_siw_oulu} shows ACER for the injection rate of the poisoned data, and the lower row shows ASR for the triggered images.
Note that when the injection rate is 0\%, no backdoor poisoning attack is performed.
For many protocols in both datasets, ACER remains low when the injection rate is less than 60\%.
This result indicates that the proposed backdoor poisoning attack can be performed without significantly degrading the overall detection accuracy, making it difficult for the system administrator to detect the attack due to the reduced detection accuracy.

We evaluate the effectiveness of backdoor poisoning attacks using ASR.
While ACER remains low, ASR increases significantly under certain conditions.
In particular, high ASRs are obtained in protocols that evaluate generalization performance against unknown spoofing attack methods.
Specifically, the evaluation against STDN using SiW exhibits high ASR in Protocol 3 (Train: Print, Test: Display), and ASR reaches almost 100\% at 50\% injection rate when using Hanawa et al.'s method and TIP-IM.
At this time, ACER shows almost no change, indicating a high probability that the attack will remain undetected.

We observe that the resistance to attacks varies depending on the spoofing attack detection methods.
In Protocol 1 of SiW, STDN failed to improve ASR while maintaining ACER, whereas PatchNet succeeded in increasing ASR to almost 100\%.
This result indicates that detection methods affect vulnerability to backdoor poisoning attacks.

The relationship between the quality of the poisoned data and ASR is shown in Fig. \ref{fig:lpips_vs_asr}.
Learned Perceptual Image Patch Similarity (LPIPS) \cite{lpips} is an image quality evaluation metric that is close to human perception, with smaller values indicating higher image quality.
While LGC and TIP-IM exhibit significant degradation in image quality, the proposed method demonstrates the possibility of performing poisoning while maintaining high image quality on both datasets.
Furthermore, while there is a general tendency for ASR to increase with larger LPIPS, the proposed method achieves both high image quality and high ASR.

The above experimental results demonstrate that the proposed backdoor poisoning attack can cause the trigger image to be falsely detected as a ``Live'' image and can be a realistic threat.

\section{Conclusion}

This paper proposed a backdoor poisoning attack method against face spoofing attack detection methods.
The proposed method utilized face feature embedding, a technique from face de-identification, to generate poisoned data.
By replacing a portion of the training data with the poisoned data, we enable a backdoor attack that prevents the detection of specific spoofing attacks.
Our experimental evaluation on public datasets demonstrates that backdoor poisoning poses a practical threat to face spoofing attack detection methods.

\printbibliography

\end{document}